\documentclass[10pt]{article} 

\usepackage[accepted]{rlj}           

\usepackage{amssymb}            
\usepackage{mathtools}          
\usepackage{mathrsfs}           
\usepackage{graphicx}           
\usepackage{subcaption}         
\usepackage[space]{grffile}     
\usepackage{url}                
\usepackage{lipsum}             

\usepackage{float}
\usepackage{multirow}
\usepackage{tabularx}
\usepackage{threeparttable}
\usepackage{booktabs}

\usepackage{xspace}
\usepackage[textsize=tiny]{todonotes}
\usepackage{hyperref} 
\usepackage{cleveref}
\newcommand{\algo}[1]{\textsc{#1}}
\newcommand{\method}{\mbox{\algo{FB-MEBE}}\xspace}

\def\gA{{\mathcal{A}}}

\def\gD{{\mathcal{D}}}

\def\gM{{\mathcal{M}}}

\def\gS{{\mathcal{S}}}

\def\gZ{{\mathcal{Z}}}

\def\sR{{\mathbb{R}}}

\DeclareMathOperator*{\argmax}{arg\,max}
\DeclareMathOperator*{\argmin}{arg\,min}

\definecolor{ourblue}{rgb}{0.368,0.507,0.71}
\definecolor{ourorange}{rgb}{0.881,0.611,0.142}
\definecolor{ourgreen}{rgb}{0.56,0.692,0.195}
\definecolor{ourred}{rgb}{0.923,0.386,0.209}
\definecolor{ourdarkred}{rgb}{0.7373, 0.3765, 0.2549}
\usepackage{dashrule}

\usepackage{algorithm, algorithmic}

\title{Maximum Entropy Behavior Exploration for \\ Sim2Real Zero-Shot Reinforcement Learning} 

\setrunningtitle{Maximum Entropy Behavior Exploration for Sim2Real Zero-Shot Reinforcement Learning}

\author{Jiajun Hu\textsuperscript{1,$\dagger$}, Núria Armengol Urpí\textsuperscript{2,3,$\dagger$}, Jin Cheng\textsuperscript{2}, Stelian Coros\textsuperscript{2}}

\emails{jiajun.hu@epfl.ch, \ nuriaa@ethz.ch}

\affiliations{
$^{1}$\textbf{Department of Mechanical Engineering, EPFL}\\
$^{2}$\textbf{Department of Computer Science, ETH Zurich}\\
$^{3}$\textbf{Max Planck Institute for Intelligent Systems}
\par 
$^\dagger$ Corresponding authors
}

\contribution{
We identify ineffective exploration as a primary bottleneck when deploying online Forward-Backward (FB) representations on real quadrupedal robots, characterized by poor state coverage and degenerate locomotion behaviors.
}
{
Prior work \citep{touati2023does} has demonstrated the effectiveness of FB in simulated control benchmarks, but its exploration properties in real-world robotic systems have not been systematically examined.
}

\contribution{
We introduce an entropy maximization sampling strategy for online exploration, improving exploration without relying on external motion capture priors.
}
{
Existing real-robot FB \citep{li2025bfm} approaches depend on motion capture datasets, which limit applicability when such priors are unavailable.
}

\contribution{
To the best of our knowledge, we demonstrate the first deployment of online FB on a real quadrupedal robot without external dataset or motion capture data.
}
{
Previous deployments of zero-shot reinforcement learning methods on physical robots have taken place on humanoid robots, for which the availability of large-scale motion capture datasets facilitate exploration.
}


\summary{Zero-shot reinforcement learning aims to learn policies that can adapt to arbitrary reward functions without retraining. Forward-Backward (FB) representations achieve this by learning successor measures, enabling efficient reward inference at test time. However, in the absence of external datasets, we observe that undirected exploration in FB often suffers from ineffective exploration, leading to distributional shrinkage and poor coverage of dynamically diverse states. In this work, we investigate the exploration bottleneck of \textit{online} FB training for quadrupedal control. We show that the empirical training data distribution progressively collapses toward low-diversity regions, significantly degrading zero-shot performance. To address this, we propose \method, a method for unsupervised behavior exploration,  that promotes exploration towards maximizing the entropy of the achieved behavior distribution. Our method actively expands state coverage without altering the FB objective. Empirical results demonstrate improved data diversity, enhanced zero-shot performance, and stable locomotion behaviors in simulated downstream tasks, and seamless transfer of recovered policies to hardware.  
}

\begin{document}

\makeCover  
\maketitle  

\begin{abstract}
Zero-shot reinforcement learning (RL) algorithms aim to learn a family of policies from a reward-free dataset, and recover optimal policies for any reward function directly at test time. Naturally, the quality of the pretraining dataset determines the performance of the recovered policies across tasks. However, pre-collecting a relevant, diverse dataset without prior knowledge of the downstream tasks of interest remains a challenge. 
In this work, we study \textit{online} zero-shot RL for quadrupedal control on real robotic systems, building upon the Forward-Backward (FB) algorithm. We observe that undirected exploration yields low-diversity data, leading to poor downstream performance and rendering policies impractical for direct hardware deployment. Therefore, we introduce ~\method, an online zero-shot RL algorithm that combines an unsupervised behavior exploration strategy with a regularization critic. \method promotes exploration by maximizing the entropy of the achieved behavior distribution. Additionally, a regularization critic shapes the recovered policies toward more natural and physically plausible behaviors. 
We empirically demonstrate that \method achieves improved performance compared to other exploration strategies in a range of simulated downstream tasks, and that it renders natural policies that can be seamlessly deployed to hardware without further finetuning. Videos and code available on our \href{https://math-286-pro.github.io/FB-MEBE-Web/}{website}.

\begin{figure}[H]
    \centering
    \includegraphics[width=1.0\linewidth]{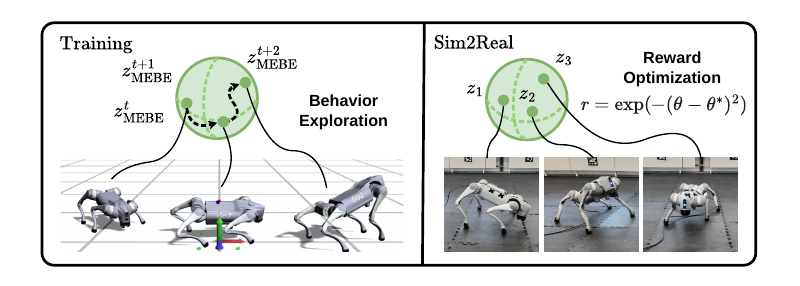} 
    \caption{\method improves behavior exploration by maximizing the behavior entropy during training. With the regularizer critic, \method achieves zero-shot hardware deployment by optimizing reward at inference time.}
    \label{fig:teaser}
\end{figure}
\end{abstract}

\section{Introduction}
Zero-shot reinforcement learning (RL) \citep{touati2023does} has emerged as a promising paradigm for developing general-purpose agents. 
By learning successor measures that decouple dynamics from rewards, these methods generate near-optimal policies for a big family of reward functions at test time, with minimal amount of computation and relying solely on unlabeled, precollected data. 
This departs from standard, fixed-reward-based RL \citep{sutton1998introduction}, which typically requires training a new policy whenever the objective changes. 
As these methods scale beyond complex and broader settings, they are usually referred to as Behavior Foundation Models (BFMs) \citep{tirinzoni2025zero}. Such generalization capabilities are particularly appealing for robotics, where usually learning relevant policies involves tedious reward engineering. While general-purpose foundation models trained on vast amounts of (un)labeled data have emerged as a promising paradigm for robotic manipulation \citep{intelligence2025pi, zitkovich2023rt,kim2025openvla}, the application of BFMs for legged systems still remains underexplored.
One of the main bottlenecks is the reliance on large, diverse, task-agnostic precollected datasets \citep{laskin2021urlb}, which are often unavailable at scale for legged robots. Recently, \citet{tirinzoni2025zero} proposed an online zero-shot RL algorithm grounding policy learning and exploration towards an external motion capture dataset. Based on this idea, BFM-Zero \citep{li2025bfm} represents an early effort to transfer such zero-shot RL methods to a humanoid robot. However, these methods still rely on supervision from external datasets. 
In this paper, we focus on the quadrupedal setting. We observe that naive deployment of online FB suffers from ineffective exploration and has difficulties in discovering the sparse manifold of stable locomotion. In contrast to previously explored efforts \citep{li2025bfm}, large motion capture datasets are not readily available for quadrupedal robots and thus cannot be used to regularize exploration.

We thus propose \method (Maximum Entropy Behavior Exploration), an online zero-shot RL framework based on the forward-backward (FB) algorithm \citep{touati2021learning}, which does not require access to external datasets.

\method leverages an unsupervised behavior exploration strategy that guides the agent to maximize the entropy of the achieved behavior distribution,  similarly to \citep{pitis2020maximum}. 
Closest to our work, \citep{urpiepistemically} framed the exploration problem in FB as a minimization of the epistemic uncertainty that the algorithm has of its representations.
In our case, we adopt a frequentist approach, guiding the agent towards reaching behaviors with low visitation count. This strategy implicitly induces a curriculum on the behaviors with which to explore next, effectively pushing the frontier of achievable behaviors. Furthermore, to ensure that the manifold of solutions induced by the unsupervised strategy is aligned with the downstream task of interest,  we embed an inductive bias to favor physically plausible behaviors via a behavior regularizer. This forces the recovered policies to maintain their zero-shot capabilities while being readily deployable to real hardware. 

We empirically demonstrate that \method achieves improved downstream performance compared to other exploration strategies in simulation in a range of downstream tasks, and that it renders policies with natural behaviors that can be seamlessly deployed to hardware without further finetuning.
To the best of our knowledge, this is the second instantiation of the FB algorithm on real robotic systems, and the very first to do so entirely online without relying on prior external datasets.
In summary, in this work we provide a maximum entropy exploration strategy for efficiently learning FB representations online that 
a) exhibits improved downstream performance in quadruped tasks in simulation compared to other exploration alternatives,
b) renders policies with smooth, physically plausible behaviors and 
c) enables the direct, zero-shot deployment of recovered policies onto real hardware.

\section{Related Work}
\paragraph{Zero-shot RL} Zero-shot RL spans a range of methods that train agents on unsupervised data to enable zero-shot generalization to a wide range of tasks. It traces back to successor representations \citep{dayan1993improving}, a direct extension to the continuous MDPs being the successor features \citep{barreto2017successor}. In \citep{borsauniversal}, given a handcrafted feature map on the states, it defines the set of all reward functions that lie on the linear span of the features. Several extensions have been proposed to learn the features map \citep{hansenfast, laskin2021urlb}. The forward backward algorithm, FB \citep{touati2021learning, touati2023does}, applies a similar idea to a finite-rank model of successor measures, adopting a contrastive loss computing pairwise dot-products across each training batch, and learns the feature map and corresponding successor features \textit{jointly}. FB has been subsequently extended to alternative parameterizations \citep{bagatella2025td, cetinfiner}, for zero-shot imitation \citep{pirotta2024fast}, offline training from low quality data \citep{jeen2024zero}, online training regularized with unlabeled expert data\citep{tirinzoni2025zero}, training on environments with different dynamics \citep{bobrin2025zero}, extending it to optimising general utilities beyond linear RL \citep{bagatella2026soft}  online finetuning \citep{sikchifast}, test-time task inference \citep{rupf2025optimistic}  or self-supervised exploration \citep{urpiepistemically, sun2025unsupervised}.
Other zero-shot methods are HILP \citep{park2024foundation}, that learn state features with a goal reaching loss in a latent space and PSM \citep{agarwal2025proto} learning an affine decomposition of the successor measure for a discrete codebook of policies.

\paragraph{Exploration in unsupervised RL}

There has been vast work on exploration in unsupervised RL \citep{eysenbachdiversity, burda2018exploration, emergentoffdads, park_metra_2024} or unsupervised goal-conditioned RL focused on discovering a wide range of goals and learning corresponding goal-reaching policies \citep{bagaria2019option,mendonca2021discovering}, exploring with goals that have lower density \citep{,florensa2018automatic,pitis2020maximum, pong2020skew, diaz2026discover}. Most of these works, in contrast to zero-shot RL require separate methods for learning policies. 
In the zero-shot realm, VISR \citep{hansenfast} and its successor APS \citep{liu2021aps} learn features online, where the representation is directly optimized to satisfy a diversity criterion. \method also learns features online but decouples the representation-learning objective from the exploration objective. Exploration in \method is driven by setting intrinsic goals that maximize the entropy of the achieved behavior distribution. Closest to our work, \citet{urpiepistemically} framed the exploration problem in online FB as a minimization of epistemic uncertainty, exploring with task embeddings that lead to the highest information gain, while \citet{sun2025unsupervised} leveraged RND \citep{burda2018exploration} to promote data diversity.

\section{Background}

We consider a standard reward-free Markov decision process (MDP), defined by the tuple \mbox{$\gM = (\gS, \gA, P, \gamma, d_0)$}, where $\gS, \gA$ are the state and action space, $P$ is the transition kernel, $\gamma \in (0,1)$ is the discount factor and $d_0$ is the initial state distribution.
Given the MDP $\gM$, a policy \mbox{$\pi: \gS \rightarrow \Delta\gA$}, mapping states to probabilities over actions, its successor measure \citep{dayan1993improving, blier2021learning} for any initial state-action pair \mbox{$(s_0, a_0) \in \gS \times \gA$} is the discounted cumulative distribution of visiting future states when starting from state $s_0 \in \gS$, take action $a \in \gA$ and following policy $\pi$ thereafter. Formally, it is defined as
\begin{equation}
\label{eq:succesor_measure}
    M^\pi( X | s_0, a_0 ) = \sum_{t\geq0}\gamma^t P(s_{t+1} \in X | s_0, a_0, \pi) \quad \forall X \in \gS.
\end{equation}
Conveniently, the action-value function for policy $\pi$ and any reward function $r: \gS \rightarrow \sR$, can be expressed as
\begin{equation}
\label{eq:q_from_successor}
Q_r^\pi(s,a) := \mathbb{E} \big[ \sum_{t\geq0}\gamma^t r(s_{t+1}) | s,a,\pi \big] = \int_{s'\in\gS} M^\pi(ds' \mid s,a)\, r(s'),
\end{equation}
which decomposes the Q-function between a reward-agnostic measure $M^\pi$, modeling the evolution of the policy in the environment, and the reward function.
There exist several algorithms for learning zero-shot policy agents \citep{eysenbachc, borsauniversal, agarwal2025proto}, but in this work we will focus on the Forward Backward (FB) \citep{touati2021learning,touati2023does} algorithm. See related work and \Cref{appendix:extended_rel_work} for extensive analysis.

Given a representation space $\gZ \subseteq \sR^d$ (usually the unit hypersphere of radius $\sqrt{d}$, and a family of policies $\pi_{z\in\gZ}$, parameterized by $z$, the FB algorithm aims to learn a \textit{forward} representation $F^\pi : \mathcal{S} \times \mathcal{A} \times \mathcal{Z} \to \mathcal{Z}$ and a \textit{backward} representation $B : \mathcal{S} \to \mathcal{Z}$
such that the successor measure in \Cref{eq:succesor_measure} admits the low-rank factorization
\begin{equation}
\label{eq:fb_factorization}
M^{\pi_z}(ds' \mid s,a) \approx  F(s,a,z)^\top B(s')\rho(ds') \quad \pi_z(s) = \argmax_a F(s,a,z)^\top z,
\end{equation}
where $\rho$ is the state distribution, $z \in \gZ$ is the task embedding of policy $\pi_z$.
The FB networks are trained to minimize a temporal-difference \textit{contrastive} loss over transitions $\{(s,a,s')\}$, independent future states $\{s^+\}$, and reward embeddings $z\sim \nu$, where $\nu$ is a distribution over $\gZ$ specified later:
\begin{align}
\label{eq:fb_loss}
\mathcal{L}_{\text{FB}}(F,B)
&=
\mathbb{E}_{\substack{
z\sim\nu,\,
(s,a,s')\sim\rho,\\
s^+\sim\rho,\,
a'\sim\pi_z(s')
}}
\Big[
\big(
F(s,a,z)^\top B(s^+)
-
\gamma
\overline{F}(s',a',z)^\top
\overline{B}(s^+)
\big)^2
\Big] \notag \\
&\quad
-
2\,\mathbb{E}_{z\sim\nu,\,(s,a,s')\sim\rho}
\Big[
F(s,a,z)^\top B(s')
\Big],
\end{align}
where $\overline{F}$ and $\overline{B}$ denote target networks and $\rho$ is the dataset distribution. In practice, as per \citet{touati2023does} additional regularization terms are used, see \Cref{app:fb_algo_details} for details. 

The $\argmax$ in \Cref{eq:fb_factorization} is approximated by jointly training an actor network with the following loss:
\begin{equation}
\label{eq:actor_loss}
\mathcal{L}_{\text{actor}}(\pi)=-\mathbb{E}_{z\sim\nu,\; s\sim\rho,\; a\sim\pi_z(s)}
\left[
F(s,a,z)^\top z
\right].
\end{equation}

After training, given a reward function $r$ specified at test time, we can find the task embedding $z_r$ parameterizing the optimal learnt policy for reward $r$, by performing a simple linear regression of $r$ onto the span of the learnt features $B$. For FB, the reward embedding can be computed simply via $z_r = \mathbb{E}_{s\sim\rho}\big[B(s)r(s)\big]$ \citep{touati2021learning}, where the expectation is approximated through sampling from the dataset distribution used for training (or a subset).

If \Cref{eq:fb_factorization} holds, then for any reward function $r$ in the linear span of the features, the policy $\pi_{z_r}$ is guaranteed to be optimal with optimal Q-function $Q^*_r(s,a) = F(s,a,z_r)^\top z_r$ \citep{touati2021learning}. In practice, the quality of the learnt representation depends on the chosen factorization $d$ and the coverage of the data $\rho$ used to train them.  In this work, we focus on the second issue.

The optimality criteria of FB \citep{touati2021learning} shows that it is sufficient to sample latents $z$ from a distribution $\nu$ that has support over the entire hypersphere. Previous works \citep{tirinzoni2025zero} propose biasing a fraction $r$ of the latent samples for exploration and training to allocate more model capacity for goal-reaching tasks. Our work focuses on a novel sampling strategy for efficient exploration.\footnote{Remark: Most of the existing works leveraging the FB framework \citep{touati2023does, cetinfiner, pirotta2024fast} are offline methods, where $\rho$ is the dataset distribution collected by decoupled unsupervised exploration methods. Our work is in the \textit{online} setting instead, where we do not have access to external datasets and we collect data transitions jointly with the learning of the FB algorithm.}

\section{Limitations of Regularized Undirected Exploration }
\label{sec:limitations}
Naturally, how well we can recover the optimal policies depends on the quality of data used to learn them.  However, when $d$ is finite and when the dataset has narrow data distributions, the learning objective lacks sufficient inductive bias on which policies to favor. This often leads to a collapse of the FB representations.

\begin{figure}[h]
    \centering
    \includegraphics[width=0.7\linewidth]{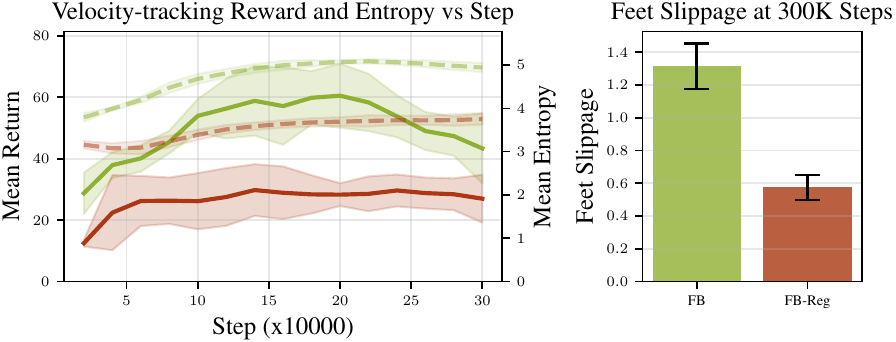}

    \centering
    \minipage{\linewidth}
    \small
    \centering
    \textcolor{ourgreen}{\rule[2pt]{20pt}{3pt}} \textrm{FB} \quad
    \textcolor{ourdarkred}{\rule[2pt]{20pt}{3pt}} \textrm{\algo{FB-Reg}}\quad
    \endminipage\
    \caption{Limitations of undirected exploration: Standard FB algorithm with undirected exploration (FB) suffers from performance degradation and entropy stagnation (left, higher is better) and yields unnatural behaviors (right, lower is better). While introducing a behavior regularizer (FB-Reg) leads to more plausible behaviors (right), it leads to severe performance degradation and further reduces behavior diversity (left).}
    \label{fig:motivation_method}
\end{figure}

 Consistent with empirical observations in \citep{tirinzoni2025zero}, our findings demonstrate this failure mode. In \Cref{fig:motivation_method} we report the average zero-shot performance when using an \textit{undirected} exploration strategy, i.e., exploring by uniformly sampling random reward embeddings from the hypersphere. We observe that this strategy 
suffers from performance degradation and entropy stagnation. Additionally, the recovered policies exhibit severe foot dragging which prevents seamless sim-to-real transfer. To mitigate these issues, we propose to ground the unsupervised policy learning by adding a regularizer. Formally, this leads to the actor loss objective:
\begin{equation}
\mathcal{L}_{\text{FB-Reg}}(\pi)
=
-\mathbb{E}_{z\sim\nu,\; s\sim\rho,\; a\sim\pi_z(s)}\Big[
F(s,a,z)^\top z
+
\lambda_{\text{reg}}\,
Q_{\text{reg}}(s,a)
\Big]
\label{eq:actor_loss_with_reg}
\end{equation}
where $Q_\text{reg}$ is a critic trained with a behavior regularizer reward. We refer interested readers to the details at \Cref{appendix:regularization_reward,appendix:qregu}. While this modification regularizes the solution manifold by effectively suppressing foot dragging (see \Cref{fig:motivation_method} (right)), we find that exploration degrades further, resulting in lower entropy and performance. We hypothesize that this is because the behavior-regularized critic assigns higher $Q_\text{reg}$ values when the robot remains stationary, which further biases the policy toward conservative behaviors and suppresses FB exploration.
To tackle these issues, in the next section we introduce \method, an algorithmic modification to FB exploration strategies to improve downstream task performance.

\section{Maximum Entropy Behavior Exploration}
\subsection{Density-Inverse Exploration for Online FB}

\textit{How should a zero-shot RL agent explore in an environment in order to accumulate diverse, relevant data, while aligned with downstream task of interest?} While we lack a proper characterization of which data is relevant \textit{a priori}, the support of the empirical training distribution $\rho$ must cover the occupancy measures of policies optimal for downstream tasks $z^d_r$. Hence, an exploration strategy should select intrinsic behaviors $z_r^t$ to bring the agent's historical achieved behavior distribution $p^t_{z^a_r}$ close to the downstream behavior distribution $p_{z^d_r}$. This can be formalized with the following distribution matching objective:
\begin{equation}
\label{eq:oracle_loss}
    J_{\text{oracle}}(p^t_{z^a_r}) = D_{KL}(p_{z^d_r} \vert \vert p^t_{z^a_r})
\end{equation}
where $p^t_{z^a_r}$ is the distribution of achieved behaviors in the replay buffer at time $t$ and $D_{KL}$ is the forward KL divergence (support covering).
Unfortunately, in the unsupervised setting, objective in \Cref{eq:oracle_loss} is intractable, because we do not have access to the desired behavior distribution $p_{z^d_r}$. In the absence of a useful inductive bias on the direction where to expand the support of $p_{z^a_r}$, similarly to \citet{pitis2020maximum}, we expand the support on all directions maximizing the entropy of the achieved behavior distribution $H[p^t_{z^a_r}]$. This leads to the following objective:
\begin{equation}
\label{eq:entropy_loss}
    J_{\text{MEBE}}(p^t_{z^a_r}) = -H[p^t_{z^a_r}],
\end{equation}
implicitly inducing a curriculum on the behaviors with which explore next to effectively push the frontier of achievable behaviors. 
To optimize \Cref{eq:entropy_loss}, at time $t$ we act with an exploration policy $\pi_t^E$ 
with reward embedding $z_r^{E}$ such that
\begin{equation}
\label{eq:z_mebe}
    z_r^{E} = \argmax_{z_{r} \in \gZ}\mathbb{E}_{z'_r \sim q(z'_r \vert z_r)} H\big[p^t_{z^a_r \cup z'_r}\big].
\end{equation} 
Importantly, \Cref{eq:z_mebe} depends on $q(z'_r \vert z_r)$ which encodes the conditional probability of achieving behavior $z_r'$ if we act with the policy parameterized with $z_r$, which we do not know. 
However, since the uniform distribution over a bounded support maximizes entropy,  increasing $H[p^t_{z_r^a}]$ encourages an even allocation of probability mass across the behavior space.

Consequently, rather than predicting future outcomes, \method steers the achieved behavior distribution towards uniformity by sampling exploration behaviors $z_r^E$ that have the minimum density in the current buffer.

\begin{figure}[h]
    \centering
    \includegraphics[width=0.9\linewidth]{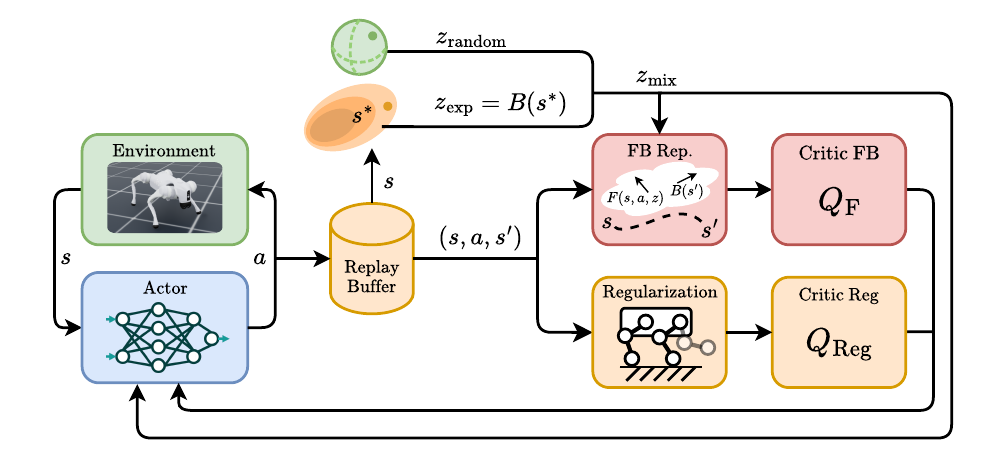}
    \caption{Maximum Entropy Behavior Forward-Backward Exploration:  \method is an online zero-shot RL algorithm that collects data by acting with behaviors $z_{exp}$ that maximize the entropy of the achieved behavior distribution. Collected data is used to train policies with a regularized loss combining policy improvement objective based on the FB action value function ($Q_{FB}$) and a critic ($Q_{reg}$) trained on a behavior regularizer reward to induce meaningful locomotion patterns.}
    \label{fig:fig_method}
\end{figure}

\subsection{Practical FB-MEBE}
Let $q_\psi(z^a_r)$ be the density model trained on the achieved behaviors stored in the replay buffer. To encourage exploration toward under-visited behaviors, we sample exploration behaviors according to a distribution that is inversely proportional to their estimated density, i.e.,
\begin{equation}
\label{eq:sample_inv}
   z_r^E \propto ({q}_\psi(z^a_r) + \epsilon)^{-\beta}
\end{equation}
where $\epsilon > 0$ ensures numerical stability and $\beta \ge 0$ controls the strength of the inverse reweighting.

In practice, because we have direct access to states, we learn the density model on the space of achieved states $q_\psi(s)$ instead (or on a projection $\varphi(s)$\footnote{To focus exploration on relevant parts of the state space, we learn the density on projections of it i.e., $\varphi: S \rightarrow S'$, where $S\subseteq\sR^d, S' \subseteq \sR^{m}, m<d$.} of it). This effectively focuses on goal-reaching behaviors, i.e., $z_r=B(s)$, and could circumvent challenges arising from representation drifts that occur due to $B$ representations changing through time. 

 Various approaches can be used to estimate the state density, such as Kernel Density Estimation (KDE), Variational Autoencoders (VAE) \citep{kingma2013auto}, k-NN particle estimators, Random Network Distillation (RND) \citep{burda2018exploration} or normalizing flows \citep{ghugare2025normalizing}. We chose normalizing flows (NFs) to fit our density $q_\psi(s)$ over all the alternatives due to its efficient likelihood evaluation, compared to quadratic computation cost of non-parametric models, while offering true density estimates compared to surrogates produced by RND or VAE. Leveraging NFs for unsupervised goal sampling has also been explored in \citep{ghugare2025normalizing}. Note that during the inverse sampling process, the state $s$ is sampled from the replay buffer rather than from the flow model to improve training stability. Details for the NF can be found in \Cref{tab:nf_hyperparameters}. 

Finally, we explore with the set of policies with reward embedding
\begin{equation}
\label{eq:practical_fb_mebe}
\{z_j^E = B(s_j^E)\}_{j=1}^K,  \quad s_j^E \sim ({q}^t_\psi(s))^{-\beta}
\end{equation}

where $K$ is the number of exploration policies collecting data in parallel and ${q}^t_\psi(s)$ is the density model fitted on data from the replay buffer at episode $t$.

Finally, as reported in \Cref{sec:limitations}, \method regularizes exploration by enforcing relevant gait patterns via a behavior regularizer, as in \Cref{eq:actor_loss_with_reg}. 
We provide the pseudocode for our algorithm in \Cref{algo:fb-mebe} and a visual schematic of \method in \Cref{fig:fig_method}. For details, see \Cref{appendix:hyperparams,appendix:qregu}.

\begin{algorithm}
\caption{Maximum entropy behavior exploration (\method)}\label{algo:fb-mebe}
\begin{algorithmic}[1]
    \STATE \textbf{Input:} $F_{\theta}$ and $\bar{F}_{\theta}$, $B_\phi$ and $\bar{B}_{\phi}$, $q_\psi$, $Q_\mathrm{reg}$ 
    \WHILE{not converged}
        \STATE Sample behavior $z_r^E$ according to \cref{eq:practical_fb_mebe}.
        \STATE Collect data $\gD_n =\mathrm{Rollout}(\pi(z_r^E))$.
        \STATE Add data to buffer $\gD_{1:n} = \gD_{1:n-1} \cup \gD_n$.
        \STATE Fit $F_\theta$, $B_\phi$, policies $\pi_z$, $q_\psi$ and $Q_\mathrm{reg}$  with $\gD_{1:n}$.
    \ENDWHILE
\end{algorithmic}
\end{algorithm}
\vspace{-2mm}

\section{Experiments}
Our experiments are designed to address the following questions:
\begin{enumerate}
    \renewcommand{\labelenumi}{\roman{enumi})}
    \item Does \method produce more natural behaviors than unregularized counterparts?
    \item Does \method achieve gains in performance comparing with other exploration strategies? 
    \item Does \method enable effective transfer from simulation to real-world hardware?
\end{enumerate}
\paragraph{Environment.} All experiments are conducted in IsaacLab simulation environment~\citep{mittal2025isaac}. We use the Unitree Go2 quadruped robot with $12$ degrees of freedom. The action space is defined as $\mathcal{A} \subseteq [-1, +1]^{12}$. The simulation runs at $200$\,Hz, while the PD controller operates at $50$\,Hz. Details about state space and domain randomization can be found in \Cref{appendix:observation-space}.

\paragraph{Task Metrics.}
We evaluate \method on two different sets of downstream tasks: velocity-tracking tasks and base orientation tasks. Both the velocity-tracking and base-orientation settings consists of $17$ downstream tasks. Performance is measured by the average episode return achieved on each task, with the episode length equals 250 steps and a maximum achievable reward of $250$. Detailed downstream task reward are in appendix \Cref{appendix:reward-function}. We additionally report the foot sliding to assess locomotion quality (details in \Cref{appendix:regularization_reward}).

\label{experiments-baselines}
\paragraph{Baselines.} We compare \method with several baselines and ablations. FB \citep{touati2021learning} performs undirected exploration by uniformly sampling random reward embeddings from the hypersphere. FB-Reg uses the same uniform exploration strategy as FB but includes a behavior regularizer (see \Cref{sec:limitations}). This isolates the performance gains provided by our exploration strategy from the gains provided by the regularizer. \method ($\beta$=2, our method) includes a behavior regularizer and directs exploration sampling 80\% of its embeddings with our proposed MEBE strategy (\Cref{eq:practical_fb_mebe}) and 20\% using uniform random embeddings. Finally, we include an ablation FB-MEBE ($\beta$=0), which sets the reverse sampling coefficient $\beta$ to zero. This can be seen as an instantiation of FB-CPR \citep{tirinzoni2025zero}, with no data prior. Instead of prioritizing rare behaviors, it performs uniform exploration over the achieved behaviors.  This isolates the performance gains provided by our exploration strategy from the gains provided by the biasing towards goal-reaching behaviors. For other ablations on the $\beta$ parameter and on sampling strategies, see \Cref{appendix:ablation_beta,appendix:ablation_exptrain}, respectively.

In practice, to focus exploration on relevant parts of the state space, we conduct exploration over a lower-dimensional projections of the state rather than directly operating on the full state space. Formally we define a projection mapping $\varphi: S \rightarrow S'$, where $S\subseteq\sR^d, S' \subseteq \sR^{m}, m<d$. For velocity-tracking tasks, exploration is performed in the planar base velocity space $\left[v_x, v_y\right]$. For base-orientation tasks exploration is performed in the space of the gravity vector expressed in the robot base frame $\left[g_x, g_y, g_z\right]$, which characterizes the base orientation. 
While this restricts the exploration space to task-relevant settings, \method could be easily extended to operate directly on the full state-space. However, its performance is bound by the accuracy of density estimators, which inherently degrade in high-dimensional spaces.

\subsection{Main Results}
\begin{figure}[h]
    \centering
    \begin{minipage}{\textwidth}
    \centering
    \includegraphics[width=0.8\linewidth]{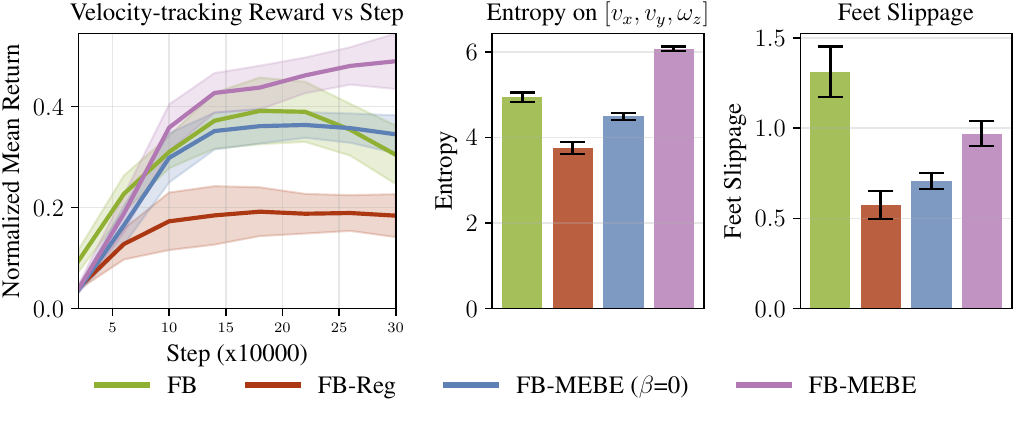}
    \end{minipage}

    \begin{minipage}{\textwidth}
    \centering
    \includegraphics[width=0.8\linewidth]{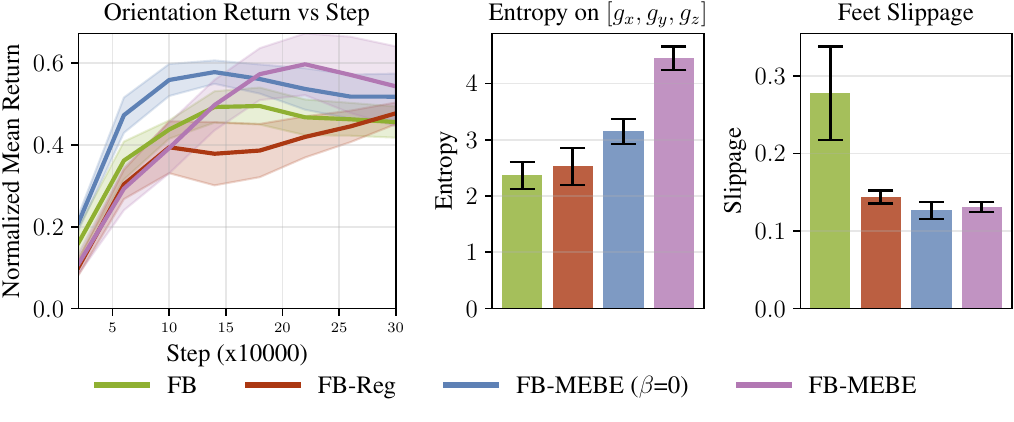}
    \end{minipage}
    \caption{
    Left: Zero-shot performance averaged over 17 downstream velocity tracking tasks (top) and orientation tasks (bottom). We plot normalized return wrt the mean return of Fast-TD3 \citep{seo2025fasttd3}. 
    Results are averaged across 5 random seeds, with shaded regions indicating $\pm$ 1-standard deviation. 
    Middle: Policy entropy on $\left[v_x, v_y, w_z \right]$ (top)  and $\left[g_x, g_y, g_z\right]$ (bottom) measured at 300K training steps.
    Right: Feet slippage measured at 300K training steps (lower is better).}
    \label{fig:FB-MEBE}
\end{figure}

\begin{figure}[h]
    \centering
    \includegraphics[width=1.0\linewidth]{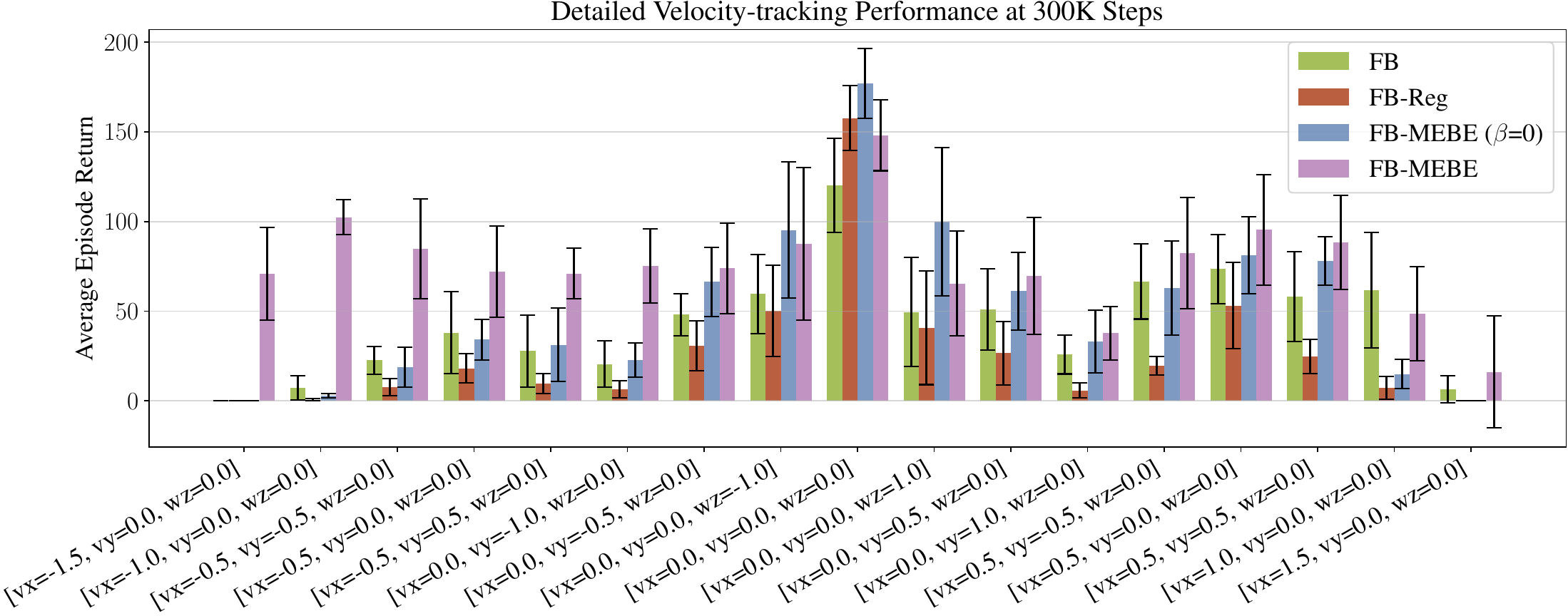}
    \caption{
    Zero-shot scores for locomotion tasks.
    Results are averaged over 5 random seeds, and error bars denote $\pm$ 1-standard deviation.
    }
    \label{fig:FB-MEBE-Detailed-Perforance}
\end{figure}

\begin{figure}[h]
    \centering
    \includegraphics[width=1.0\linewidth]{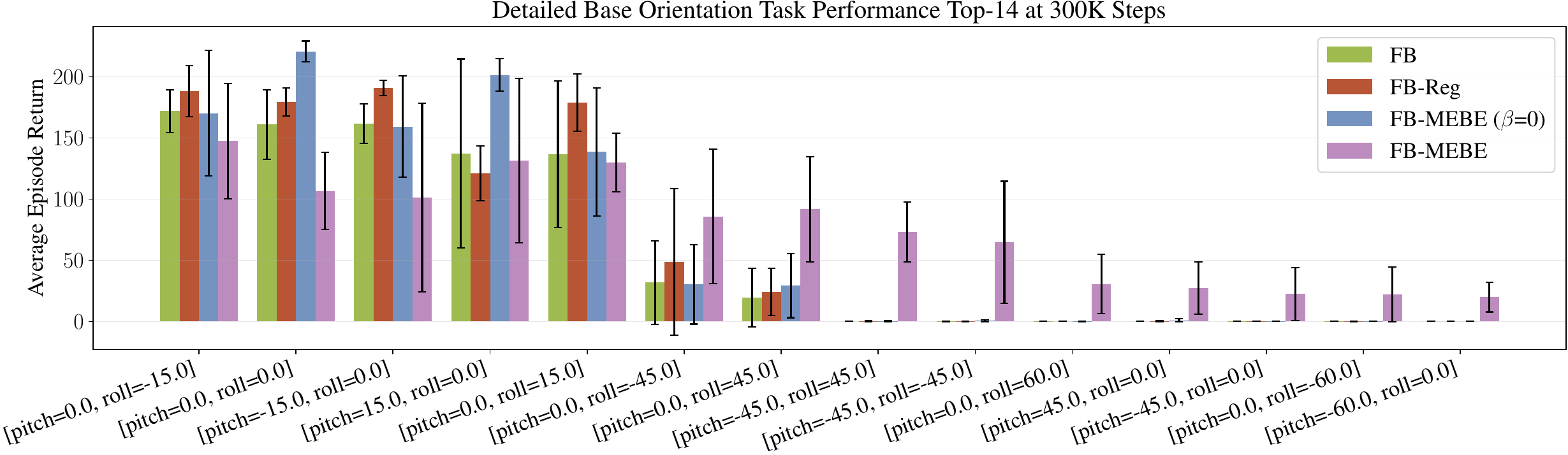}
    \caption{
    Zero-shot scores for orientation tasks. Results are averaged over 5 random seeds, and error bars denote $\pm$ 1-standard deviation.
    }    
    \label{fig:orientation}
\vspace{-2mm}
\end{figure}

\paragraph{Overall Performance.}
In \Cref{fig:FB-MEBE}, we report results on several downstream tasks. All returns are normalized using the mean return achieved by Fast-TD3 \citep{seo2025fasttd3}. Fast-TD3 is a supervised reinforcement learning algorithm trained directly with task-specific rewards, and therefore serves as an approximate upper bound under explicit reward supervision. In contrast, all FB-based methods are trained without access to downstream task rewards and must rely on unsupervised exploration to acquire useful behaviors. 
 \method achieves an average performance that is the highest over all the other baselines for velocity tracking tasks or on par with best performing baseline for orientation tasks. Most notably, it yields the highest entropy among all baselines. This indicates that \method expands behavioral coverage during online training. Importantly, this broader exploration translates into improved downstream task performance rather than unstructured behavior diversification. This can be further seen in \Cref{fig:FB-MEBE-Detailed-Perforance,fig:orientation} where \method outperforms all other baselines in extreme downstream tasks. For orientation tasks, \method shows lower sample efficiency than \method-$\beta=0$, showcasing that unprioritized goal-reaching exploration can be a valid strategy for some downstream tasks. In contrast, FB and FB-Reg perform the worst among all algorithms, which we attribute to uninformed exploration (FB) and the additional regularization constraints (FB-Reg) that overly restrict exploration, especially in velocity-tracking tasks, leading to limited behavioral coverage and consequently lower average performance.  Regarding gait-related metrics, FB-MEBE substantially reduces the feet slide penalty compared to the original FB, indicating that the regularization successfully suppresses unnatural dragging behaviors. Although FB-MEBE exhibits slightly higher feet slippage than FB-Reg, this difference is expected. Since FB-MEBE actively explores higher locomotion velocities, increased horizontal foot velocities naturally lead to higher sliding penalties. Therefore, the observed difference reflects expanded behavioral coverage rather than degraded gait stability. 

\paragraph{Detailed Locomotion Performance.}
\Cref{fig:FB-MEBE-Detailed-Perforance} reports performance across 17 velocity-tracking tasks. 
While all methods perform reasonably on moderate velocity commands, the baselines suffer significant degradation on high-velocity tasks. This is consistent with the replay buffer distribution (\Cref{appendix:replay_buffer_data_distribution}), where the FB baseline shows a low coverage of high-velocity states. By contrast, FB-MEBE expands the buffer coverage into sparsely visited velocity regions, maintaining strong performance even on high-velocity tasks (see \Cref{appendix:reward-function} for details).

\paragraph{Detailed Orientation Performance.}
\Cref{fig:orientation} shows the performance across 12 base orientation tracking settings. 
Notably, \method demonstrates superior performance on challenging large-angle tasks compared to baselines. This indicates enhanced exploration within the orientation behavior space.
However, we observe that FB-MEBE achieves lower returns than baselines on small-angle orientation tasks, including FB-Reg. We hypothesize that this is because small-angle commands correspond to behaviors near the initial state, which are naturally well-covered by random exploration. \method prioritizes less-visited behaviors, allocating more capacity toward exploring extreme orientations, which can reduce its performance on these more common states. FB-Reg does not exhibit this degradation because its regularization primarily constrains foot-slippage behaviors, thereby leaving orientation exploration unaffected.

\begin{figure}[H]
    \centering
    \begin{minipage}[b]{0.2\linewidth}
        \centering
        \includegraphics[width=\linewidth]{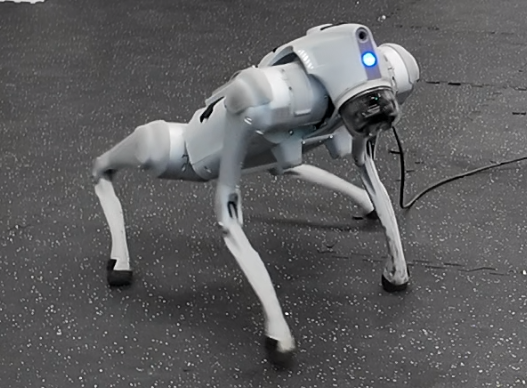}
    \end{minipage}
    \begin{minipage}[b]{0.2\linewidth}
        \centering
        \includegraphics[width=\linewidth]{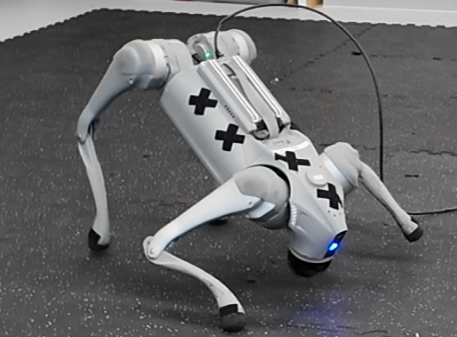}
    \end{minipage}
    \begin{minipage}[b]{0.2\linewidth}
        \centering
        \includegraphics[width=\linewidth]{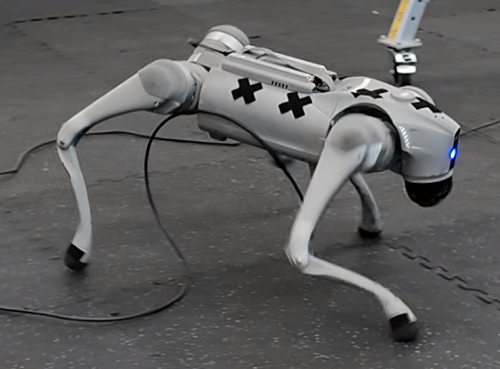}
    \end{minipage}
    \caption{\method allows for zero-shot sim2real transfer to the Unitree Go2 quadrupedal robot. }
    \label{fig:sim2real}
\end{figure}

\paragraph{Hardware Tests.}
We deploy FB-MEBE on a real Unitree Go2 robot and evaluate it on both locomotion and orientation control tasks. To facilitate real-world control, commands are issued through a joystick interface, and the robot is guided by a composite reward function (details in \Cref{appendix:reward-function}) that simultaneously regulates velocity, base orientation, and base height. As illustrated in \Cref{fig:sim2real}, FB-MEBE enables the robot to track commanded translational velocities, adjust pitch and roll orientations, and regulate its height. The learned policy also supports compositional commands, such as walking forward while maintaining a pitch angle of $15^\circ$, or tracking yaw angular velocity while keeping the base height at $0.32\,\mathrm{m}$. Importantly, the resulting behaviors remain stable and natural, without exhibiting excessive action rates or noticeable foot slippage that could destabilize the robot.

\section{Conclusion}
We propose \method, a practical online zero-shot RL algorithm that enables efficient exploration for learning quadrupedal control. \method addresses the undirected exploration of naive online FB algorithm, by using an exploration strategy that guides data collection policies towards maximizing the entropy of the achieved behavior distribution. To constrain exploration, \method introduces a behavior regularizer, shaping the recovered policies towards more natural and physically plausible patterns. Through this novel yet simple approach, \method yields policies that can be seamlessly deployed to hardware without further finetuning. To the best of our knowledge, this is the first instantiation of a fully online zero-shot RL algorithm on real robotics systems without relying on external datasets as motion priors.

While \method represents a principled attempt toward efficient online zero-shot RL, some limitations remain. The recovered policies will struggle with downstream tasks requiring behaviors that fall far outside the collected dataset. Additionally, the exploration strategy is still loosely constrained, which can occasionally lead the agent to explore extreme state configurations that are irrelevant for downstream task of interest. A deeper theoretical understanding of the FB algorithm's components could provide better intuition regarding whether FB implicitly prioritizes some behaviors over others. Finally, while we leveraged a useful behavior regularizer, developing more theoretically sound ways to characterize a specific family of reward functions of interest could lead to safer, more directed exploration in future work.

\subsubsection*{Broader Impact Statement}
\label{sec:broaderImpact}
Our work focuses on zero-shot RL algorithms for quadrupedal control. While we recognize that autonomous robotic systems have the potential to be deployed for harmful applications, our contributions are fundamental in nature and do not directly target any specific negative use cases.

\appendix





\subsubsection*{Acknowledgments}
\label{sec:ack}
The authors would like to thank Marco Bagatella and Thomas Rupf for their help in reviewing this manuscript, and the Max Planck ETH Center for Learning Systems for supporting Núria Armengol.


\bibliography{main}

@book{sutton1998introduction,
    title={Reinforcement Learning: {A}n Introduction},
    author={Sutton, Richard S. and Barto, Andrew G.},
	publisher={The MIT Press},
	year={1998},
	address={Cambridge, MA},
}

@article{li2025bfm,
  title={Bfm-zero: A promptable behavioral foundation model for humanoid control using unsupervised reinforcement learning},
  author={Li, Yitang and Luo, Zhengyi and Zhang, Tonghe and Dai, Cunxi and Kanervisto, Anssi and Tirinzoni, Andrea and Weng, Haoyang and Kitani, Kris and Guzek, Mateusz and Touati, Ahmed and others},
  journal={arXiv preprint arXiv:2511.04131},
  year={2025}
}

@article{seo2025fasttd3,
  title={Fasttd3: Simple, fast, and capable reinforcement learning for humanoid control},
  author={Seo, Younggyo and Sferrazza, Carmelo and Geng, Haoran and Nauman, Michal and Yin, Zhao-Heng and Abbeel, Pieter},
  journal={arXiv preprint arXiv:2505.22642},
  year={2025}
}

@inproceedings{sun2025unsupervised,
  title={Unsupervised zero-shot reinforcement learning via dual-value forward-backward representation},
  author={Sun, Jingbo and Tu, Songjun and Zhang, Qichao and Li, Haoran and Liu, Xin and Chen, Yaran and Chen, Ke and Zhao, Dongbin},
  booktitle={International Conference on Learning Representations},
  volume={2025},
  pages={25604--25633},
  year={2025}
}

@inproceedings{eysenbachdiversity,
  title={Diversity is All You Need: Learning Skills without a Reward Function},
  author={Eysenbach, Benjamin and Gupta, Abhishek and Ibarz, Julian and Levine, Sergey},
  booktitle={International Conference on Learning Representations},
  year={2018}
}

@inproceedings{pong2020skew,
  title={Skew-Fit: State-Covering Self-Supervised Reinforcement Learning},
  author={Pong, Vitchyr and Dalal, Murtaza and Lin, Steven and Nair, Ashvin and Bahl, Shikhar and Levine, Sergey},
  booktitle={International Conference on Machine Learning},
  pages={7783--7792},
  year={2020},
  organization={PMLR}
}

@article{touati2021learning,
  title={Learning one representation to optimize all rewards},
  author={Touati, Ahmed and Ollivier, Yann},
  journal={Advances in Neural Information Processing Systems},
  volume={34},
  pages={13--23},
  year={2021}
}

@article{dayan1993improving,
  title={Improving generalization for temporal difference learning: The successor representation},
  author={Dayan, Peter},
  journal={Neural computation},
  volume={5},
  number={4},
  pages={613--624},
  year={1993},
  publisher={MIT Press}
}

@article{barreto2017successor,
  title={Successor features for transfer in reinforcement learning},
  author={Barreto, Andr{\'e} and Dabney, Will and Munos, R{\'e}mi and Hunt, Jonathan J and Schaul, Tom and van Hasselt, Hado P and Silver, David},
  journal={Advances in neural information processing systems},
  volume={30},
  year={2017}
}

@inproceedings{liu2021aps,
  title={Aps: Active pretraining with successor features},
  author={Liu, Hao and Abbeel, Pieter},
  booktitle={International Conference on Machine Learning},
  pages={6736--6747},
  year={2021},
  organization={PMLR}
}

@article{burda2018exploration,
  title={Exploration by random network distillation},
  author={Burda, Yuri and Edwards, Harrison and Storkey, Amos and Klimov, Oleg},
  journal={arXiv preprint arXiv:1810.12894},
  year={2018}
}

@inproceedings{touati2023does,
  title={Does Zero-Shot Reinforcement Learning Exist?},
  author={Touati, Ahmed and Rapin, J{\'e}r{\'e}my and Ollivier, Yann},
  booktitle={The Eleventh International Conference on Learning Representations},
  year={2023}
}

@article{laskin2021urlb,
  title={Urlb: Unsupervised reinforcement learning benchmark},
  author={Laskin, Michael and Yarats, Denis and Liu, Hao and Lee, Kimin and Zhan, Albert and Lu, Kevin and Cang, Catherine and Pinto, Lerrel and Abbeel, Pieter},
  journal={arXiv preprint arXiv:2110.15191},
  year={2021}
}

@inproceedings{eysenbachc,
  title={C-Learning: Learning to Achieve Goals via Recursive Classification},
  author={Eysenbach, Benjamin and Salakhutdinov, Ruslan and Levine, Sergey},
  booktitle={International Conference on Learning Representations},
  year={2021}
}

@article{blier2021learning,
  title={Learning successor states and goal-dependent values: A mathematical viewpoint},
  author={Blier, L{\'e}onard and Tallec, Corentin and Ollivier, Yann},
  journal={arXiv preprint arXiv:2101.07123},
  year={2021}
}

@article{jeen2024zero,
  title={Zero-shot reinforcement learning from low quality data},
  author={Jeen, Scott and Bewley, Tom and Cullen, Jonathan M},
  journal={Advances in Neural Information Processing Systems},
  volume={37},
  pages={16894--16942},
  year={2024}
}

@inproceedings{hansenfast,
  title={Fast Task Inference with Variational Intrinsic Successor Features},
  author={Hansen, Steven and Dabney, Will and Barreto, Andre and Warde-Farley, David and Van de Wiele, Tom and Mnih, Volodymyr},
  booktitle={International Conference on Learning Representations},
  year={2019}
}

@inproceedings{pirotta2024fast,
  title={Fast imitation via behavior foundation models},
  author={Pirotta, Matteo and Tirinzoni, Andrea and Touati, Ahmed and Lazaric, Alessandro and Ollivier, Yann},
  booktitle={International Conference on Learning Representations},
  volume={2024},
  pages={12685--12724},
  year={2024}
}

@article{tirinzoni2025zero,
  title={Zero-shot whole-body humanoid control via behavioral foundation models},
  author={Tirinzoni, Andrea and Touati, Ahmed and Farebrother, Jesse and Guzek, Mateusz and Kanervisto, Anssi and Xu, Yingchen and Lazaric, Alessandro and Pirotta, Matteo},
  journal={arXiv preprint arXiv:2504.11054},
  year={2025}
}

@inproceedings{pitis2020maximum,
  title={Maximum entropy gain exploration for long horizon multi-goal reinforcement learning},
  author={Pitis, Silviu and Chan, Harris and Zhao, Stephen and Stadie, Bradly and Ba, Jimmy},
  booktitle={International Conference on Machine Learning},
  pages={7750--7761},
  year={2020},
  organization={PMLR}
}

@inproceedings{bagaria2019option,
  title={Option discovery using deep skill chaining},
  author={Bagaria, Akhil and Konidaris, George},
  booktitle={International Conference on Learning Representations},
  year={2019}
}

@inproceedings{sikchifast,
  title={Fast Adaptation with Behavioral Foundation Models},
  author={Sikchi, Harshit and Tirinzoni, Andrea and Touati, Ahmed and Xu, Yingchen and Kanervisto, Anssi and Niekum, Scott and Zhang, Amy and Lazaric, Alessandro and Pirotta, Matteo},
  booktitle={Reinforcement Learning Conference},
  year={2025}
}

@article{bobrin2025zero,
  title={Zero-shot adaptation of behavioral foundation models to unseen dynamics},
  author={Bobrin, Maksim and Zisman, Ilya and Nikulin, Alexander and Kurenkov, Vladislav and Dylov, Dmitry},
  journal={arXiv preprint arXiv:2505.13150},
  year={2025}
}

@inproceedings{borsauniversal,
  title={Universal Successor Features Approximators},
  author={Borsa, Diana and Barreto, Andre and Quan, John and Mankowitz, Daniel J and van Hasselt, Hado and Munos, Remi and Silver, David and Schaul, Tom},
  booktitle={International Conference on Learning Representations},
  year={2019}
}

@inproceedings{urpiepistemically,
  title={Epistemically-guided forward-backward exploration},
  author={Urp{\'\i}, N{\'u}ria Armengol and Vlastelica, Marin and Martius, Georg and Coros, Stelian},
  booktitle={Reinforcement Learning Conference 2025},
  year={2025}
}

@article{rupf2025optimistic,
  title={Optimistic task inference for behavior foundation models},
  author={Rupf, Thomas and Bagatella, Marco and Vlastelica, Marin and Krause, Andreas},
  journal={arXiv preprint arXiv:2510.20264},
  year={2025}
}

@article{bagatella2026soft,
  title={Soft Forward-Backward Representations for Zero-shot Reinforcement Learning with General Utilities},
  author={Bagatella, Marco and Rupf, Thomas and Martius, Georg and Krause, Andreas},
  journal={arXiv preprint arXiv:2602.06769},
  year={2026}
}

@inproceedings{park2024foundation,
  title={Foundation Policies with Hilbert Representations},
  author={Park, Seohong and Kreiman, Tobias and Levine, Sergey},
  booktitle={International Conference on Machine Learning},
  pages={39737--39761},
  year={2024},
  organization={PMLR}
}

@inproceedings{agarwal2025proto,
  title={Proto Successor Measure: Representing the Behavior Space of an RL Agent},
  author={Agarwal, Siddhant and Sikchi, Harshit and Stone, Peter and Zhang, Amy},
  booktitle={International Conference on Machine Learning},
  pages={566--586},
  year={2025},
  organization={PMLR}
}

@inproceedings{cetinfiner,
  title={Finer Behavioral Foundation Models via Auto-Regressive Features and Advantage Weighting},
  author={Cetin, Edoardo and Touati, Ahmed and Ollivier, Yann},
  booktitle={Reinforcement Learning Conference},
  year={2025}
}

@article{bagatella2025td,
  title={Td-jepa: Latent-predictive representations for zero-shot reinforcement learning},
  author={Bagatella, Marco and Pirotta, Matteo and Touati, Ahmed and Lazaric, Alessandro and Tirinzoni, Andrea},
  journal={arXiv preprint arXiv:2510.00739},
  year={2025}
}

@misc{park_metra_2024,
	title = {{METRA}: {Scalable} {Unsupervised} {RL} with {Metric}-{Aware} {Abstraction}},
	shorttitle = {{METRA}},
	doi = {10.48550/arXiv.2310.08887},
	urldate = {2024-11-17},
	publisher = {arXiv},
	author = {Park, Seohong and Rybkin, Oleh and Levine, Sergey},
	month = mar,
	year = {2024},
	note = {arXiv:2310.08887},
}

@article{emergentoffdads,
  author       = {Archit Sharma and
                  Michael Ahn and
                  Sergey Levine and
                  Vikash Kumar and
                  Karol Hausman and
                  Shixiang Gu},
  title        = {Emergent Real-World Robotic Skills via Unsupervised Off-Policy Reinforcement
                  Learning},
  journal      = {CoRR},
  volume       = {abs/2004.12974},
  year         = {2020},
  eprinttype    = {arXiv},
  eprint       = {2004.12974},
  bibsource    = {dblp computer science bibliography, https://dblp.org}
}

@article{mendonca2021discovering,
  title={Discovering and achieving goals via world models},
  author={Mendonca, Russell and Rybkin, Oleh and Daniilidis, Kostas and Hafner, Danijar and Pathak, Deepak},
  journal={Advances in Neural Information Processing Systems},
  volume={34},
  pages={24379--24391},
  year={2021}
}

@article{diaz2026discover,
  title={Discover: Automated curricula for sparse-reward reinforcement learning},
  author={Diaz-Bone, Leander and Bagatella, Marco and H{\"u}botter, Jonas and Krause, Andreas},
  journal={Advances in Neural Information Processing Systems},
  volume={38},
  pages={21863--21896},
  year={2026}
}

@article{kingma2013auto,
  title={Auto-encoding variational bayes},
  author={Kingma, Diederik P and Welling, Max},
  journal={arXiv preprint arXiv:1312.6114},
  year={2013}
}

@article{ghugare2025normalizing,
  title={Normalizing flows are capable models for rl},
  author={Ghugare, Raj and Eysenbach, Benjamin},
  journal={arXiv preprint arXiv:2505.23527},
  year={2025}
}

@article{intelligence2025pi,
  title={$\pi^{*}_{0.6}$: a VLA That Learns From Experience},
  author={Intelligence, Physical and Amin, Ali and Aniceto, Raichelle and Balakrishna, Ashwin and Black, Kevin and Conley, Ken and Connors, Grace and Darpinian, James and Dhabalia, Karan and DiCarlo, Jared and others},
  journal={arXiv preprint arXiv:2511.14759},
  year={2025}
}

@inproceedings{zitkovich2023rt,
  title={Rt-2: Vision-language-action models transfer web knowledge to robotic control},
  author={Zitkovich, Brianna and Yu, Tianhe and Xu, Sichun and Xu, Peng and Xiao, Ted and Xia, Fei and Wu, Jialin and Wohlhart, Paul and Welker, Stefan and Wahid, Ayzaan and others},
  booktitle={Conference on Robot Learning},
  pages={2165--2183},
  year={2023},
  organization={PMLR}
}

@inproceedings{kim2025openvla,
  title={OpenVLA: An Open-Source Vision-Language-Action Model},
  author={Kim, Moo Jin and Pertsch, Karl and Karamcheti, Siddharth and Xiao, Ted and Balakrishna, Ashwin and Nair, Suraj and Rafailov, Rafael and Foster, Ethan P and Sanketi, Pannag R and Vuong, Quan and others},
  booktitle={Conference on Robot Learning},
  pages={2679--2713},
  year={2025},
  organization={PMLR}
}

@inproceedings{florensa2018automatic,
  title={Automatic goal generation for reinforcement learning agents},
  author={Florensa, Carlos and Held, David and Geng, Xinyang and Abbeel, Pieter},
  booktitle={International conference on machine learning},
  pages={1515--1528},
  year={2018},
  organization={PMLR}
}

@article{mittal2025isaac,
  title={Isaac lab: A gpu-accelerated simulation framework for multi-modal robot learning},
  author={Mittal, Mayank and Roth, Pascal and Tigue, James and Richard, Antoine and Zhang, Octi and Du, Peter and Serrano-Munoz, Antonio and Yao, Xinjie and Zurbr{\"u}gg, Ren{\'e} and Rudin, Nikita and others},
  journal={arXiv preprint arXiv:2511.04831},
  year={2025}
}
\bibliographystyle{rlj}

\beginSupplementaryMaterials

\section{Algorithm Details}
\label{app:fb_algo_details}
Following prior work on Forward-Backward (FB) representations \citep{touati2023does}, we include two additional regularization losses to stabilize FB training. First, we adopt an orthonormality regularization on the backward representation $
\Sigma_B = \mathbb{E}\left[B(s)B(s)^\top\right]$,
which encourages the covariance of $B(s)$ to approach the identity matrix. This regularization prevents the backward representation from collapsing and improves the conditioning of the learned embedding space. Second, we introduce a temporal-difference regularization loss on the forward map that encourages $
F(s,a,z)^\top z
$ to approximate the action-value function corresponding to the reward induced by $B(s)^\top \Sigma_B^{-1}z$. This loss encourages the forward representation to capture directions in the latent space that are relevant for policy optimization. See \citep{touati2023does} (Appendix B) for more details.

\section{Practical Exploration}
\label{appendix:practical-exploration}
To improve exploration efficiency in locomotion exploration, we discard states where the robot pitch or roll exceeds $21.5^\circ$ (note that in orientation exploration we keep these states, since we want robot to reach extreme behavior). These states typically correspond to failure events such as falling, which can produce large velocities unrelated to meaningful locomotion behaviors. Filtering such degenerate transitions helps focus exploration on valid locomotion regimes.


\section{Ablation Tests}
\subsection{Ablation Study on $\beta$.} 
\label{appendix:ablation_beta}
\begin{figure}[H]
    \centering
    \includegraphics[width=0.6\linewidth]{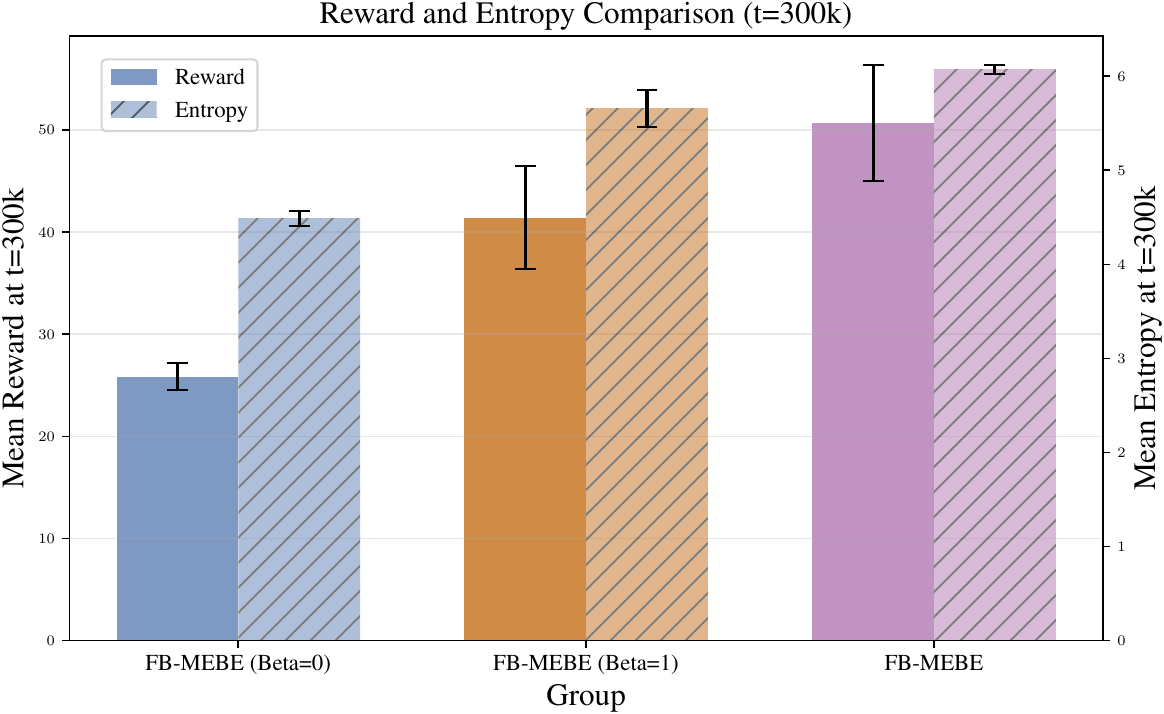}
    \caption{Comparison between different betas on FB-MEBE on locomotion tasks}
    \label{fig:Abalation}
\end{figure}

\Cref{fig:Abalation} presents the ablation results with different inverse-density coefficients $\beta$. When $\beta=0$, corresponding to uniform latent sampling without density reweighting, both entropy and task reward remain relatively low. As $\beta$ increases, entropy and reward consistently improve.

These results confirm that inverse-density reweighting effectively increases the entropy of the achieved behavior distribution and mitigates replay buffer occupancy collapse during online training.

\subsection{Ablation study on sampling strategy}
\label{appendix:ablation_exptrain}
\begin{figure}
    \centering
    \includegraphics[width=0.8\linewidth]{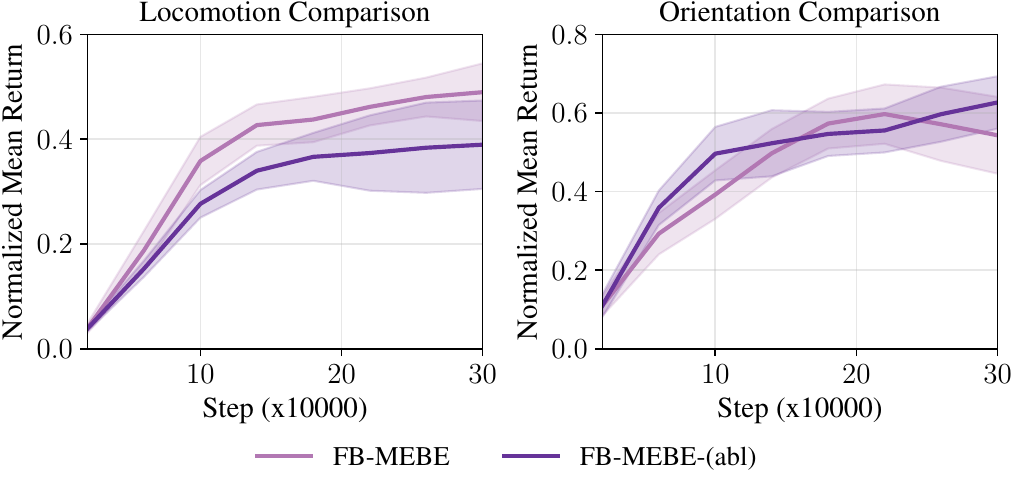}
    \caption{Comparison on velocity-tracking tasks between \method and an ablation \method-\text{abl} leveraging different sampling strategies for training the FB algorithm}
    \label{fig:ablation_exptrain1}
\end{figure}

We evaluate a version of \method (\method-\text{abl}), that performs an ablation on the sampling strategy used for \textit{training} the FB representations.
\method uses the \textit{same} sampling strategy for reward embeddings used for data collection ($0.8 z_{exp}$ + $0.2 z_{random}$) and the reward embedding used for training FB representations.
Here, we compare with an alternative version which only uses our  strategy for data collection, and uses $(0.8 z_\text{goal reaching} + 0.2 z_\text{random}$ for training. The later is analogous to the exploration strategy $\method(\beta=0)$ uses.
While \method puts more emphasis on learning FB representations and policies for low visitation behaviors and might hence learn better policies useful for exploration,\method-\text{abl} samples exploration behaviors uniformly from the replay buffer.
While these two approaches are different in nature, in practice we do not find significant differences as can be seen in \Cref{fig:ablation_exptrain1}. Extensive evaluation on different mixing ratios can also bring more insights on better sampling alternatives.

\section{Replay buffer data distribution}
\label{appendix:replay_buffer_data_distribution}
\begin{figure}[H]
    \centering
    \includegraphics[width=0.8\linewidth]{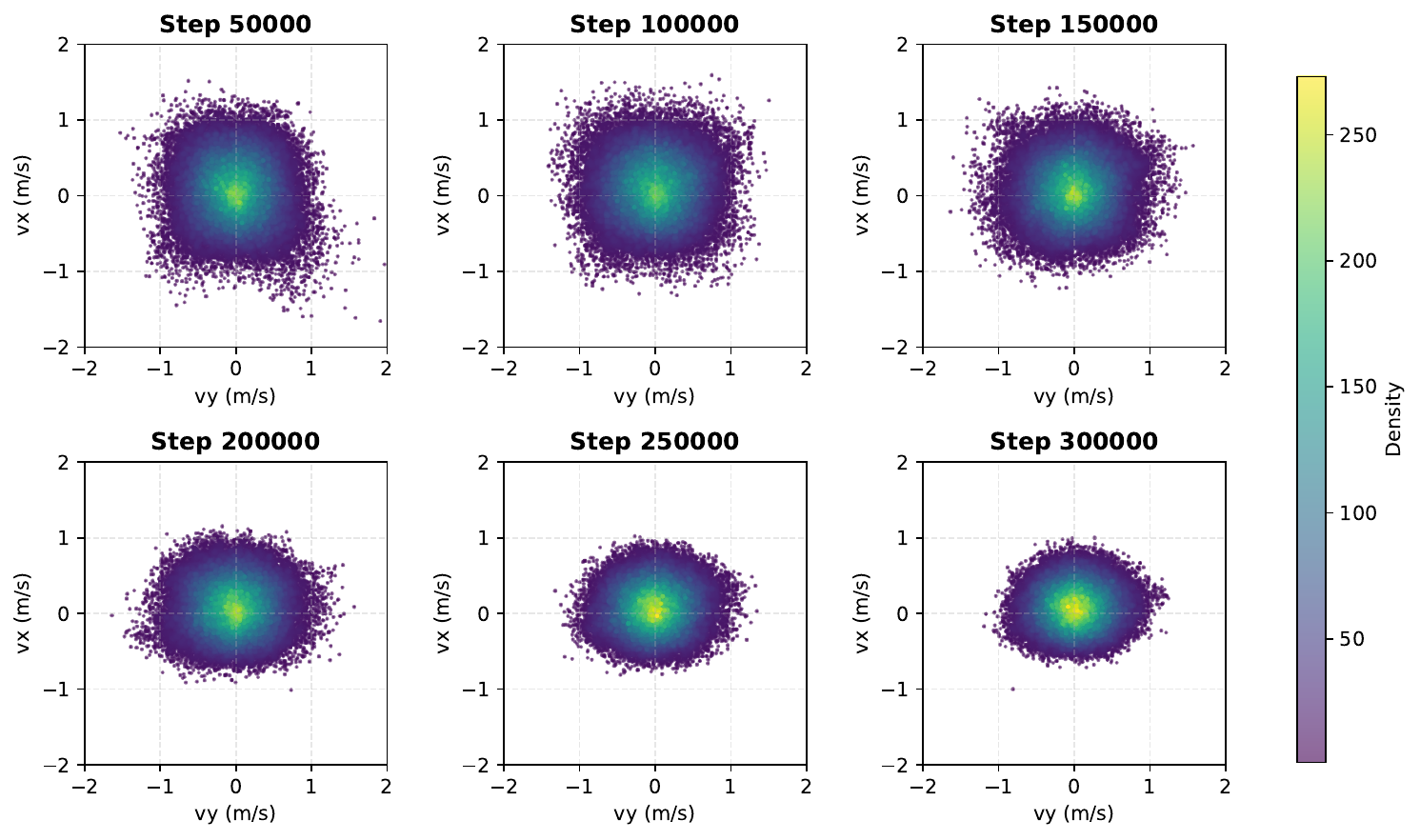}
    \caption{Velocity $\left[v_x, v_y \right]$ distribution of FB replay buffer at different training step. The color represents the density of data points.}
    \label{fig:replay_buffer_data_distribution_fb}
\end{figure}

\label{appendix:replay_buffer_data_distribution_fb_mebe}
\begin{figure}[H]
    \centering
    \includegraphics[width=0.8\linewidth]{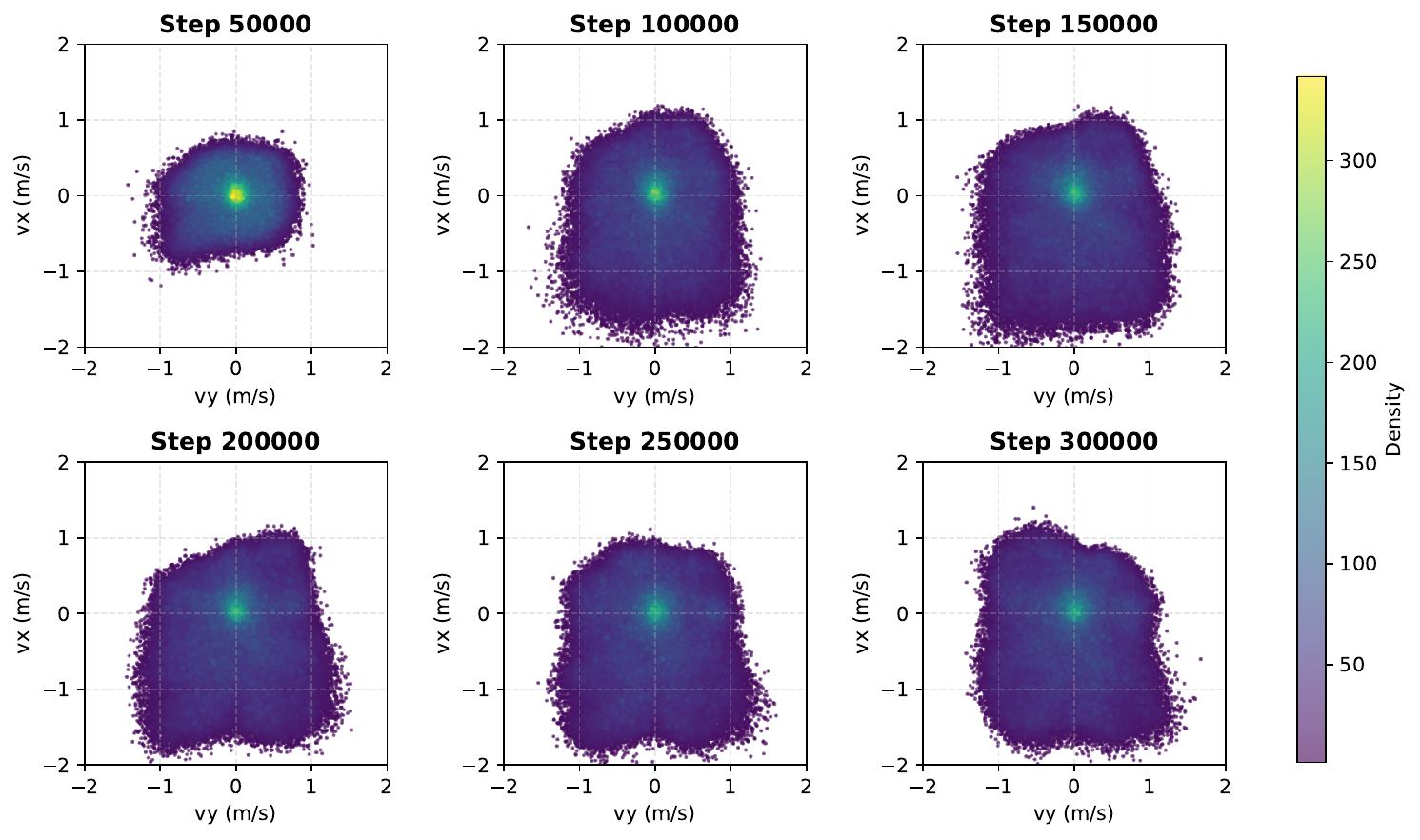}
    \caption{Velocity $\left[v_x, v_y \right]$ distribution of FB-MEBE replay buffer at different training step. The color represents the density of data points.}
    \label{fig:replay_buffer_data_distribution_fb_mebe}
\end{figure}
\section{Task Reward Function}
\label{appendix:reward-function}

 Let $\mathbf{v} \in \mathbb{R}^3$ and $\boldsymbol{\omega}_z \in \mathbb{R}$ denote the current base linear velocity and yaw angular velocity, respectively,  $h \in \mathbb{R}$ denote the current base height and let $\mathbf{g} \in \mathbb{R}^3$ denote the projected gravity vector. The corresponding command targets are denoted by $\mathbf{v}^\ast$, $\omega_z^\ast$,  $h^*$, and $\mathbf{g}^\ast$.
 
We define the tracking errors as
\begin{align}
e_v &= \lVert \mathbf{v} - \mathbf{v}^\ast \rVert_2, \\
e_\omega &= \lvert \omega_z - \omega_z^\ast \rvert, \\
e_g &= \lVert \mathbf{g} - \mathbf{g}^\ast \rVert_2 \\
e_h &=  \lVert h - h^\ast \rVert_2.
\end{align}
\\
The individual reward terms are computed using Gaussian-shaped functions:
\begin{align}
r_v &= \exp\left(-\left(\frac{e_v}{0.3}\right)^2\right), \\
r_\omega &= \exp\left(-\left(\frac{e_\omega}{0.2}\right)^2\right), \\
r_g &= \exp\left(-\left(\frac{e_g}{0.1}\right)^2\right), \\
r_h &= \exp\left(-\left(\frac{e_h}{0.05}\right)^2\right).
\end{align}

The locomotion and orientation task reward is defined as follows:

\begin{align}
r_{\text{locomotion}} &= r_v \cdot r_\omega \cdot r_g \\
r_{\text{orientation}} &= r_g \cdot r_h
\end{align}

The reward for sim-to-real transfer is defined by multiplication of all terms:
\begin{equation}
r_{\text{sim2real}} = r_v \cdot r_\omega \cdot r_g \cdot r_h
\end{equation}

\section{Regularization rewards}
\label{appendix:regularization_reward}

We provide the mathematical definitions of the regularization reward.  Let $N_j$ denote the number of actuated joints and $N_f$ the number of feet. At time step $t$, let $\ddot{\mathbf{q}}_t \in \mathbb{R}^{N_j}$ be the joint acceleration vector, $\mathbf{a}_t \in \mathbb{R}^{N_j}$ be the action vector, and $\mathbf{a}_{t-1}$ be the previous action.  For each foot $i \in \{1,\dots,N_f\}$, let $h_{t,i}$ be its vertical position (world frame), and $\mathbf{v}^{xy}_{i,t} \in \mathbb{R}^{2}$ be its horizontal linear velocity (world frame).

\paragraph{Joint acceleration penalty.}
The joint-acceleration penalty is defined as the squared $\ell_2$ norm of the joint acceleration vector:
\begin{equation}
r_{\text{joint\_acc}}
=
\|\ddot{\mathbf{q}}_t\|_2^2
=
\sum_{k=1}^{N_j} \left(\ddot{q}_{t,k}\right)^2 .
\end{equation}

\paragraph{Action rate penalty.}
The action-rate penalty discourages abrupt changes between consecutive actions:
\begin{equation}
r_{\text{action\_rate}}
=
\|\mathbf{a}_t - \mathbf{a}_{t-1}\|_2^2
=
\sum_{k=1}^{N_j} \left(a_{t,k}-a_{t-1,k}\right)^2 .
\end{equation}

\paragraph{Feet slippage.}
The feet-slide penalty penalizes horizontal foot motion during ground contact.
A binary contact indicator $c_{t,i}$ is defined as
\begin{equation}
c_{t,i}
=
\begin{cases}
1, & \text{if } h_{t,i} > h_{\mathrm{feet\_contact}}, \\
0, & \text{otherwise}.
\end{cases}
\end{equation}

\begin{equation}
r_{\mathrm{feet\_slide}}
=
\sum_{i=1}^{N_f}
c_{t,i}
\,
\|\mathbf{v}_{t,i}^{xy}\|_2 .
\end{equation}

\paragraph{Total typical regularization reward.}
In our experiments, the regularization reward used to train the critic is:
\begin{equation}
r_{\text{reg}} =
-2.5\times 10^{-7}\, r_{\text{joint\_acc}}
-0.1\, r_{\text{action\_rate}}
-0.1\, r_{\text{feet\_slide}}
\end{equation}

\section{Regularized Exploration.}
\label{appendix:qregu}
We use critic $Q_{\text{reg}}(s,a)$ to estimate the long-term return of the regularization reward $r_{\text{reg}}$ defined in \Cref{appendix:regularization_reward}. Given tuples $(s,a,r_{\text{reg}},s')$ sampled from the replay buffer, we train the regularization critic by minimizing a standard temporal-difference (TD) loss:
\begin{equation}
\mathcal{L}_{\text{critic}}(Q_\text{reg})
=
\mathbb{E}_{ (s,a,s')\sim \rho,a'\sim\pi_z (s')}\Big[
\big(
Q_{\text{reg}}(s,a)
-
\big(r_{\text{reg}}
+
\gamma \,
\bar Q_{\text{reg}}(s', a')
\big)
\big)^2
\Big]
\label{eq:reg_critic_loss}
\end{equation}
\\
Therefore, to pretrain FB agent and learn good gait at the same time, we have the following actor loss:
\begin{equation}
\mathcal{L}_{\text{actor}}(\pi)
=
-\mathbb{E}_{z\sim\nu,\; s\sim\rho,\; a\sim\pi_z(s)}\Big[
Q_{\text{FB}}(s,a,z)
+
\lambda_{\text{reg}}\,
Q_{\text{reg}}(s,a)
\Big]
\end{equation}

\newpage 
\section{Simulation Setup}

\paragraph{Joint Control}
We employ joint position control as the low-level control interface. The joint configuration of the robot in the standing pose is defined as the nominal joint position $q_{nom}$. The policy outputs joint position offsets $a$, which are scaled by a factor $s$ and added to the nominal configuration to obtain the target joint positions
\begin{equation}
q_{target} = q_{nom} + s \cdot a .
\end{equation}
The target joint positions are tracked using a PD controller
\begin{equation}
\tau = K_p (q_{target} - q) + K_d (\dot{q}_{target} - \dot{q}),
\end{equation}
where $q$ and $\dot q$ denote the current joint positions and velocities.

\begin{table}[H]
\centering

\begin{minipage}[t]{0.48\textwidth}
\centering
\caption{Joint Control Parameters}
\vspace{0pt}
\begin{tabular}{l c}
\toprule
\textbf{Parameter} & \textbf{Value} \\
\midrule
Action Scale 
    & $0.5$ \\
Proportional Gain $K_p$ 
    & $25.0$\\
Derivative Gain $K_d$ 
    & $0.5$\\
Control Frequency [Hz] 
    & $50\text{Hz}$\\
\bottomrule
\end{tabular}

\end{minipage}
\end{table}

\paragraph{Domain Randomization and Observation Noise}
\begin{table}[H]
\centering

\begin{minipage}[t]{0.48\textwidth}
\centering
\caption{Domain Randomization}
\vspace{0pt}
\begin{tabular}{l c}
\toprule
\textbf{Parameter} & \textbf{Range} \\
\midrule
Friction Coefficient 
    & $\mathcal{U}([0.5,\,1.5])$ \\
Base COM Offset [m] 
    & $\mathcal{U}([-0.05,\,0.05])$ \\
Base Mass perturbation [kg]
    & $\mathcal{U}([-2.0,\,2.0])$ \\
Link Mass Offset [m] 
    & $\mathcal{U}([-0.01,\,0.01])$ \\
Link Mass perturbation [kg]
    & $\mathcal{U}([-0.2,\,0.2])$ \\
Default Joint Pos [rad] 
    & $\mathcal{U}([-0.3,\,0.3])$ \\
\bottomrule
\end{tabular}
\end{minipage}
\hfill
\begin{minipage}[t]{0.48\textwidth}
\centering
\caption{Additive Observation Noise}
\vspace{0pt}
\begin{tabular}{l c}
\toprule
\textbf{Observation} & \textbf{Range} \\
\midrule
$v$ [m/s] 
    & $\mathcal{U}([-0.1,\,0.1])$ \\
$\omega$ [rad/s] 
    & $\mathcal{U}([-0.2,\,0.2])$ \\
$\mathrm{grav}$ 
    & $\mathcal{U}([-0.05,\,0.05])$ \\
$q$ [rad] 
    & $\mathcal{U}([-0.01,\,0.01])$ \\
$\dot{q}$ [rad/s] 
    & $\mathcal{U}([-0.15,\,0.15])$ \\
\bottomrule
\end{tabular}
\end{minipage}

\end{table}


\section{Observation Space}
\label{appendix:observation-space}
\begin{table}[H]
  \centering
  \renewcommand{\arraystretch}{1.15}
  \begin{tabular}{l l}
    \toprule
    Network & Obs Dimension Breakdown\\
    \midrule

    Critic $(F)$ 
    & $ \left[ \mathbf{v}_{3}, \boldsymbol{\omega}_{3}, \mathbf{g}_{3}, h_{\text{base},1}, \mathbf{q}_{12}, \dot{\mathbf{q}}_{12} \right]$ \\

    $B$ 
    & $ \left[ \mathbf{v}_{3}, \boldsymbol{\omega}_{3}, \mathbf{g}_{3}, h_{\text{base},1} \right]$ \\

    Actor
    & $ \left[ \mathbf{v}_{3}, \boldsymbol{\omega}_{3}, \mathbf{g}_{3}, \mathbf{q}_{12}, \dot{\mathbf{q}}_{12}, \mathbf{a}_{12} \right]$ \\

    Critic (Reg)
    & $\left[ \mathbf{v}_{3}, \boldsymbol{\omega}_{3}, \mathbf{g}_{3}, h_{\text{base},1}, \mathbf{q}_{12}, \dot{\mathbf{q}}_{12}, \mathbf{a}_{12}, h_{\text{feet},4}, F_{\text{feet},4} \right]$ \\

    \bottomrule
  \end{tabular}

  \caption{Inputs for different networks}
  \label{tab:fb_inputs}
\end{table}

    \footnotesize
    \begin{flushleft}
    The subscripts represent the dimension of this variable. $\mathbf{v}$ is the base linear velocity, $\boldsymbol{\omega}$ is the base angular velocity, $\mathbf{g}$ is the projected gravity vector, $h_{\text{base}}$ is the base height, $\mathbf{q}$ and $\dot{\mathbf{q}}$ are joint positions and joint velocities, respectively, $\mathbf{a}$ is the last action, $h_{\text{foot}}$ denotes foot heights (4 feet), and $F_{\text{feet}}$ denotes foot contact forces (4 feet). $^{*}$.
    \end{flushleft}
    \normalsize

\textbf{Note on B mapping}: We note that the B representation acts on a lower-dimensional projection of the full state space instead.  When dealing with high dimensionality environments, learning future probabilities for all states is difficult and generally requires large $d$ to accommodate for all possible rewards.
In general, we are often interested in rewards that depend not on the full state but on a projection of it.
If this is known in advance, the representation B can be trained on that part of the state only, with same theoretical guarantees (Appendix, Theorem 4 \citep{touati2021learning}). Hence, we leverage an environment-dependent feature map $\vartheta: S \rightarrow G$, and learn $B(g)$ instead of $B(s)$, where $g=\vartheta(s)$.
Importantly, rewards can be arbitrary functions of g. This was also suggested in \citep{touati2021learning}.

\section{Hyperparameters}
\label{appendix:hyperparams}

\begin{table}[H]
\centering
\small
\setlength{\tabcolsep}{12pt}
\renewcommand{\arraystretch}{1.15}
\begin{tabular}{l r}
\toprule
\textbf{Hyperparameter} & \textbf{Value} \\
\midrule
Number of training steps& 150k\\
Number of parallel environments & 2048\\
Number of rollout steps between each agent update & 10\\
Number of gradient steps per agent update & 10\\
Policy delay (TD3 actor update interval) & 2 \\
Number of initial steps with random actions & 1000\\
 Number of samples for task inference&100k\\
Replay buffer size & 4M\\
Batch size & 4096\\
Discount factor & 0.98 \\
 $z$ update step interval&100\\
 $z$ dimension $d$&50\\
 $z$ distribution $\nu$&\\
 ~~ -uniform on unit sphere&20\%\\
 ~~ -goals from online buffer&80\%\\
 Coefficient for orthonormality loss&100\\
 Coefficient for Fz-regularization loss&0.1\\
 Regularization coefficient $\lambda_\text{reg}$&20\\
\bottomrule
\end{tabular}
\caption{Training hyperparameters}
\label{tab:train_hparams}
\end{table}

\label{appendix:fb_hparams}
\begin{table}[H]
\centering
\small
\setlength{\tabcolsep}{10pt}
\renewcommand{\arraystretch}{1.15}
\begin{tabular}{lcccc}
\toprule
\textbf{Hyperparameter} &
\textbf{Critic $(F)$} &
\textbf{B} &
\textbf{Critic $(\text{Reg})$} &
\textbf{Actor} \\
\midrule
Learning Rate &
$5\cdot 10^{-5}$ &
$5\cdot 10^{-5}$ &
$5\cdot 10^{-5}$ &
$5\cdot 10^{-5}$ \\

Soft update coefficient& 
$0.01$& 
$0.01$& 
$0.005$&
$0.01$\\

Input Variables &
$(s, a, z)$ &
$(s)$ &
$(s, a)$&
$(s, z)$ \\

Output Dim &
$d$ &
$d$ &
$1$ &
$12$ \\

Embedding Hidden Units &
$1024$ &
-- &
--&
$1024$ \\

Feed Forward Hidden Layers &
$2$ &
$2$ &
$4$&
$2$ \\

Feed Forward Hidden Units &
$1024$ &
$512$ &
$1024$ &
$1024$ \\

Activations &
ReLU &
ReLU &
ReLU &
ReLU \\

Number of Parallel Networks &
$2$ &
$1$ &
$2$ &
$1$ \\
\bottomrule
\end{tabular}
\caption{Network hyperparameters}
\label{tab:fb_hparams}
\end{table}

\begin{table}[H]
\centering
\caption{Hyperparameters of the Normalizing Flow.}
\begin{tabular}{l c}
\toprule
\textbf{Hyperparameter} & \textbf{Value} \\
\midrule
Flow architecture & RealNVP-style normalizing flow \\
Number of coupling layers & 10 \\
Hidden dimension & 64 \\
Activation function & ReLU \\
Scale output activation & Tanh \\
Masking strategy & Alternating checkerboard / channel mask \\
Base distribution & Standard Gaussian $\mathcal{N}(0, I)$ \\
Learning rate & $1\cdot 10^{-3}$ \\
Batch size & 256 \\
Training epochs & 30 \\
Goal buffer size & 10k \\
\bottomrule
\end{tabular}
\label{tab:nf_hyperparameters}
\end{table}

\begin{table}[H]
\centering
\caption{Hyperparameters of reverse sampling.}
\begin{tabular}{l c}
\toprule
\textbf{Hyperparameter} & \textbf{Value} \\
\midrule
normalizing flow update frequency & 1000 steps \\
sample buffer size & 100k \\
$\epsilon$ (for stability) & 0.1 \\
$\beta$ (reverse sampling strength) & 2.0 \\
\bottomrule
\end{tabular}
\label{tab:reverse_sampling_hyperparameters}
\end{table}


\section{Connection to Universal Successor Features}
\label{appendix:extended_rel_work}
Given a feature map $\phi: \gS \rightarrow \sR^d$ that embeds states into d-dimensional features,  the successor features \citep{barreto2017successor} for policy $\pi$ are defined as $\Phi^\pi(s,a) = \mathbb{E}\big[\sum_{t\geq0}\gamma^t \phi(s_{t+1}) | s,a,\pi\big]$. Then for any reward function that can be expressed as a linear combination of the features, i.e., $r(s)= z^\top\phi(s)$, where $z \in \sR^d$ are the weights, the Q-function can be expressed as: $Q_r^\pi(s,a) = z^\top\Phi^\pi(s,a)$. 
Universal successor features (USF) \citep{borsauniversal}, goes one step further and allows not only zero-shot policy evaluation, but a generic framework for zero-shot policy optimization. This is achieved by learning a family of policies $\{\pi_z\}$ parameterized by a latent vector $z \in \sR^d$, such that for each $z$, the policy $\pi_z$ is trained to be optimal with respect to reward function $r_z:= z^\top\phi$. Formally,
\begin{equation}\label{eq:usf}
   \mathbb{E}\Big[\sum_{t\geq0}\gamma^t \phi(s_{t+1}) | s,a,\pi\Big], \quad \pi_z(s) = \argmax_{a} \Phi(s,a,z)^\top  z.
\end{equation}
There exist several instantiations of USF \citep{agarwal2025proto, park2024foundation} (see \citep{touati2023does} for an extensive list), however, they requires a separate methods for learning features $\phi$ and one for learning its successor features, usually via a TD-learning approach. In contrast, the forward backward representation \citep{touati2021learning} provides an alternative framework for zero-shot RL that jointly learns the features and its successor features, via a contrastive learning approach. 

FB representations are closely related to successor features. In fact, $F(s,a,z)$ is the successor feature of policy $\pi_z$ of the basic features $\phi(s)= \text{Cov}_{\rho}(B)^{-1}B(s)$ \citep{touati2023does}, where $\text{Cov}_{\rho}B=\mathbb{E}_{\tilde s\sim\rho}\big[B(\tilde s)B(\tilde s)^\top \big]$.

After training, given a reward function $r$ specified at test time, we can find the task embedding $z_r$ parameterizing the optimal learnt policy for reward $r$, by performing a simple linear regression of $r$ onto the span of the learnt features $B$. Namely we find the weights $z_r^*$ (coordinates of the feature basis) that minimize $z_r^* = \argmin_z \mathbb E_{s\sim\rho}\big[ \big((r(s) - z_r^\top \phi(s)\big)^2] = \text{Cov}_{\rho}(\phi)^{-1} \mathbb{E}_{\tilde s\sim\rho}\big[\phi(s)r(s)\big]$. 
For FB, the reward embedding can be computed simply via $z_r = \mathbb{E}_{s\sim\rho}\big[B(s)r(s)\big]$ \citep{touati2021learning}, where the expectation is approximated through sampling from the dataset distribution used for training (or a subset).

\end{document}